# Hybrid Primal Sketch: Combining Analogy, Qualitative Representations, and Computer Vision for Scene Understanding

**Kenneth D. Forbus**  FORBUS@NORTHWESTERN.EDU
Computer Science, Northwestern University, 2233 Tech Drive, Evanston, IL, 60208, USA

**Kezhen Chen**  KZCHEN0204@GMAIL.COM
Together.ai, San Francisco, CA USA

**Wangcheng Xu**  WANGCHENG.XU@NORTHWESTERN.EDU
**Madeline Usher**  USHER@NORTHWESTERN.EDU
Computer Science, Northwestern University, 2233 Tech Drive, Evanston, IL, 60208, USA

## Abstract

One of the purposes of perception is to bridge between sensors and conceptual understanding. Marr's *Primal Sketch* combined initial edge-finding with multiple downstream processes to capture aspects of visual perception such as grouping and stereopsis. Given the progress made in multiple areas of AI since then, we have developed a new framework inspired by Marr's work, the *Hybrid Primal Sketch*, which combines computer vision components into an ensemble to produce sketch-like entities which are then further processed by CogSketch, our model of high-level human vision, to produce both more detailed shape representations and scene representations which can be used for data-efficient learning via analogical generalization. This paper describes our theoretical framework, summarizes several previous experiments, and outlines a new experiment in progress on diagram understanding.

## 1. Introduction

Perception provides organisms with information about the world around them, organized in ways that are useful given their capabilities. For people, and human-like AI systems, this includes providing a bridge between sensory data and conceptual understanding: What objects are like, how they are arranged into scenes, and how these change over time. One of the early revolutions in computer vision was Marr's (1982) *Primal Sketch*, which proposed that edges actually formed a kind of hybrid representation, a combination of metric and symbolic information that served as a starting point for many down-stream computations, such as stereopsis and lightness. For example, stereopsis was modeled by aligning edges from pairs of images, since edges provide more stable information about the world than pixels. The primal sketch was used to construct the 2 ½ D sketch, which carved images up into regions, so that properties of surfaces could be estimated (e.g. lightness and material composition).

Computer vision has made considerable progress since that time. Deep learning has produced components for building vision systems that have revolutionary capabilities. How should these advances be used to build end to end vision systems? One approach, which is receiving the most attention, is to build end-to-end systems by training over massive datasets (e.g. Google's Gemini, Open AI's GPT4/V). These end to end systems often treat language as an end goal, e.g. describing a scene or answering questions about it, which is an over-simplification. Moreover, they require massive amounts of training and suffer from hallucinations. We think that ignoring intermediate representations may be a mistake. For example, there is converging evidence from neuroscience (e.g. Amorapanth et al. 2012) that schematic representations computed from images are used in relational processing. Moreover, we have evidence that analogical processing appears to be used in human visual reasoning (Forbus & Lovett, 2022), which suggests that relational representations provide an important intermediate form of representation. These are hybrid representations, grounded in symbolic visual elements tied to locations and properties of the image, just as in the original primal sketch.

This paper describes an alternative we have been exploring, the *hybrid primal sketch* (HPS), which builds on these hypotheses. The hybrid primal sketch combines advances in computer vision with advances in qualitative visual representation and reasoning. It uses an ensemble of low-level computations, including components constructed via deep learning, to produce visual entities called *glyphs* for subsequent analysis. These glyphs are motivated by our work on sketch understanding, where a glyph consists of digital ink plus a conceptual label. The edges that make up regions or bounding boxes produced by visual components are treated as if they were digital ink. Any recognition information is treated as information about the conceptual label for the glyph. Thus the ensemble produces a set of glyphs, which are then analyzed at a higher level by CogSketch, our sketch understanding system that is also a model of aspects of high-level human vision (Forbus et al. 2011). This includes decomposing a glyph into fine-grained edge representations to capture its shape, and computing relationships between glyphs to construct relational representations of scenes. These descriptions can be used in analogical reasoning and learning. Analogical generalization provides incremental, data-efficient learning of inspectable models, producing performance on a par with, and sometimes better than, deep learning systems.

This paper describes the hybrid primal sketch (HPS) framework and illustrates it with several results, including one line of investigation in progress. We begin by providing brief key background and related work on computer vision, qualitative representations and analogical learning. Then we describe the HPS model, including how the ensemble works, glyphs as an intermediate representation, and highlights of shape and scene analysis. Completed experiments are summarized next, e.g. on sketch recognition and Kinect video analysis. An experiment in progress on diagram understanding is described next. Finally, we discuss conclusions and future work.

## 2. Background and Related Work

Here we briefly summarize necessary background and closely related work.

### 2.1 Computer Vision

A survey of even one area of computer vision is beyond the scope of this paper. Most relevant for this work are algorithms that extract and recognize visual structure. Broadly, these can be divided into edge finders, recognizers, and segmenters. Edge finders like Canny (1986) operate over natural images finding zero crossings in intensity gradients that correspond to edges. The advantage of edge-finders is that they are (ideally) fully general-purpose, placing no limitations on the kinds of objects in the images. The disadvantage is that additional processing is required to use the edges found to propose segmentations of the image into regions and more complex structures. A special case is processing bitmaps that depict sketches (e.g. the MNIST dataset). For such images a common technique is to treat them as binary images, by down-sampling and thresholding, then using a combination of edge thinners (e.g. Zhang & Suen 1984) and tracing algorithms (e.g. Potrace (Krenski & Selinger, 2003)) to produce edges. We use such techniques here to produce vector representations of sketches depicted in bitmaps.

By contrast, recognizers, like MASK-RCNN (He et al. 2018), directly provide proposed regions and conceptual labels. However, they must be trained on the kinds of entities to be recognized, and hence are not as broadly applicable as edge finders. Nonetheless, recognizers are an extremely popular choice for building vision systems today. We have used both MASK-RCNN and Faster-RCNN (Ren et al. 2015) as object recognizers over natural images (Chen & Forbus, 2021). Our techniques only assume that some form of boundary (bounding box or mask) is provided plus a word or phrase as a conceptual label and should work with other recognizers. We note than an increasingly popular choice are open-vocabulary detectors, where one must provide the set of words corresponding to the kinds of entities sought in the image (Wu et al. 2024). We also have used OWLv2 (Minderer et al. 2023) as an input component, when top-down constraints suggest what kinds of entities should be sought (e.g. arrows).

Segmenters carve up an image into regions, roughly corresponding to entities (or pieces of them). For example, Meta's Segment Anything Model (SAM; Kirillov et al. 2023) was trained on over a billion masks across 11 million images. Downstream processes must still pull together interpretations of regions into objects, but given the smaller number of regions, this is a much easier problem than not having the segmentation.

## 2.2 Qualitative Visual Representations

Qualitative representations provide symbolic representations of continuous phenomena. In visual processing qualitative representations have been introduced for both shapes (i.e. within-object relationships) and for scenes (i.e. between-object relationships). We discuss each in turn.

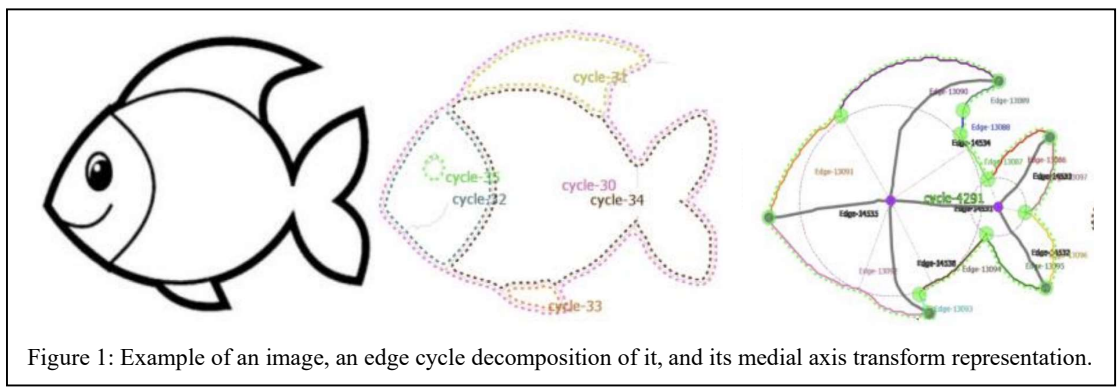

Figure 1: Example of an image, an edge cycle decomposition of it, and its medial axis transform representation.

For shapes we focus here on the special case of sketches of objects, consisting of a set of visual edges. This factors out, for example, the color of regions (but see Chen et al. 2020). Edges can be qualitatively described in terms of their shape (i.e. straight or curved, and when part of a larger contour, concave or convex), their orientation (e.g. quadrants), and relative size (e.g. within a distribution provided by a diagram) – see (Forbus et al. 2017a) for details. Edges can be combined into *edge cycles*, which provide an intermediate representation that captures aspects of surfaces (McLure et al. 2011). There is evidence that people are sensitive to the medial axis transform in shape perception (Lowet et al. 2018). Figure 1 illustrates edge cycles and the medial axis transform. These are used in the work discussed here, but there are other qualitative shape representations, e.g. (Museros & Escrig, 2004) which is used for matching for robot assembly tasks.

For scenes, a variety of qualitative representations have been used. Many are qualitative spatial calculi, which provide a set of mutually exclusive and collectively exhaustive set of relationships (e.g. Moratz, 2017). The most widely used is the eight-relation region connection calculus (RCC8, Cohn et al. 1997), which captures topological relationships like disjoint, edge-connected, overlapping, and inside. In purely qualitative spatial reasoning, transitivity tables are used to squeeze out implications of combinations of statements relating a set of objects. In vision settings like ours, such relationships can simply be computed from metric shape representations and positions of the objects. Coordinate information is also used to compute common positional relationships, e.g. above/below, left/right.

## 2.3 Sketch Understanding versus Scene Understanding

Work in artificial intelligence on sketch understanding tends to equate understanding with recognition (Forbus, 2019). In general recognition is insufficient for sketch understanding because the mapping between sketches and the objects or systems they can depict is many-to-many. When people sketch they usually talk, providing another channel of information that enables others in the interaction to understand them. In other words, even people do not rely solely on recognition when sketching with each other. Recognition is a catalyst, not a requirement. Sketching also often provides timing information, i.e. watching the ink being laid down. This extra source of information can help improve recognition, e.g. it is well known that in handwriting recognition, timing data is crucial for accuracy. In CogSketch (Forbus et al. 2011), people sketching lay down visual ink, manually segment it into entities, and provide conceptual labels from a large knowledge base. Thus errors in both recognition and segmentation are entirely factored out, at the cost of putting a little more work on the sketcher. As discussed further in Section 3, these visual entities, called *glyphs*, are proposed here as an interface layer between low-level vision and high-level vision. The more tentative nature of segmentation and recognition when building atop computer vision outputs will sometimes require probabilistic abductive interpretation, as discussed below.

A recent approach to scene understanding is to train a multimodal foundation model, e.g. LLaVA (Liu et al. 2023) & Ferret v2 (Zhang et al. 2024). However, while such problems have high performance, they also have massive hallucinations (Lu et al. 2024; Huang et al. 2024), require batch training, and produce uninspectable models. By contrast, the approach presented here incrementally produces inspectable models that are reliable.

## 2.4 Analogical Learning

We use an *analogy stack* based on Structure-Mapping Theory (Gentner, 1983). Matching is performed via the structure-mapping engine (SME, Forbus et al. 2017b), which takes as input two structured representations and produces one or more mappings between them. Each mapping indicates what statements and entities in one description correspond with the other, what inferences follow from these correspondences, and a numerical estimate of their similarity. Retrieval is performed by MAC/FAC (Forbus et al. 1995), a two-stage map reduce process. The first stage uses sparse vector dot-products over a massive library of cases to find a small number of items for the second stage, which uses SME in parallel with the structured version of the probe and the MAC output cases to select the few most similar as remindings. Generalization is performed by the Sequential Analogical Generalization Engine (SAGE; Kandaswamy & Forbus, 2012). SAGE models concepts via *generalization pools* which are incrementally constructed by examples. These pools contain both generalizations and outliers. Generalizations are formed when a new example retrieves an outlier, i.e. a prior example that wasn't similar enough to anything else so far. Each new example added to a generalization causes the probability of statements in that generalization to be updated, e.g. statements that don't show up frequently wear away. A generalization pool can have multiple generalizations, providing the ability to represent disjunctive concepts. It is similar to K-means plus outliers, but SAGE figures out the number of generalizations automatically, rather than them having to be pre-specified. There is also a working memory version of SAGE, SageWM, where the generalization pools are in working memory and retrieval is based on SME similarity scores, modulated by recency.

We note that this analogy stack supports incremental learning, but today's deep learning systems do not. Thus all the experiments used for testing are batch settings.

There have been multiple models of analogy in cognitive science (e.g. Doumas et al. 2008; Hummel & Holyoak, 1997; Mitchell 1993). Unfortunately, these models do not have the representational capacity to handle the kinds of vision tasks described here.

## 3. The Hybrid Primal Sketch

The key idea of Marr's (1982) Primal Sketch was that vision has a symbolic component. Edges were viewed as a simple primitive symbolic representation, also having continuous properties such as location and contrast, being grounded in parallel visual processing elements. These descriptions were used in a variety of downstream processes. For example, visual structure including texture would be found by performing grouping operations based on statistical properties of the edges and continuity of form constraints. Marr's group focused on both trying to understand the purpose of visual computations and how they were implemented biologically. While some of the original hypotheses of the Primal Sketch have not worked out, particularly in terms of the biological picture[1], the overall flow of operations and the construction of intermediate representations – something like edges, regions that provide clues to surfaces in the world, abductive interpretation processes to accumulate local pieces into hypotheses about the perceived world – remain of interest. That said, many efforts in computer vision carve the processing up somewhat differently. For example, recognizers trained using deep learning directly on images provide labels for some region of the image (either a bounding box or a mask approximating the outline of the object being hypothesized). Are such recognizers biologically plausible? We leave that question aside here, since our interest is more practical: How can we use the advances in computer vision to build cognitive systems that can see? Our assumption is that at some point human visual processing does produce symbolic descriptions[2], but we are agnostic as to where and how. Our concerns lie upstream: The computation of relational representations of shapes and scenes so that analogical learning can be used in visual processing. Even though visual stimuli are plentiful for most organisms, it still behooves them to learn incrementally, in a data-efficient manner. Since mistakes are inevitable in learning, organisms lucky enough to have language can discuss and refine their visual concepts, if they are inspectable. The kinds of qualitative visual and spatial representations developed in qualitative reasoning research are, we hypothesize, among the contents that can be inspected by people (and hence should be inspectable by our machines, if they are to participate in discussions of them). In other words, at the level of inspectability, we want human-like representations and processing, whereas for computations below that level, as long as they produce functionally reasonable representations, that is all that matters.

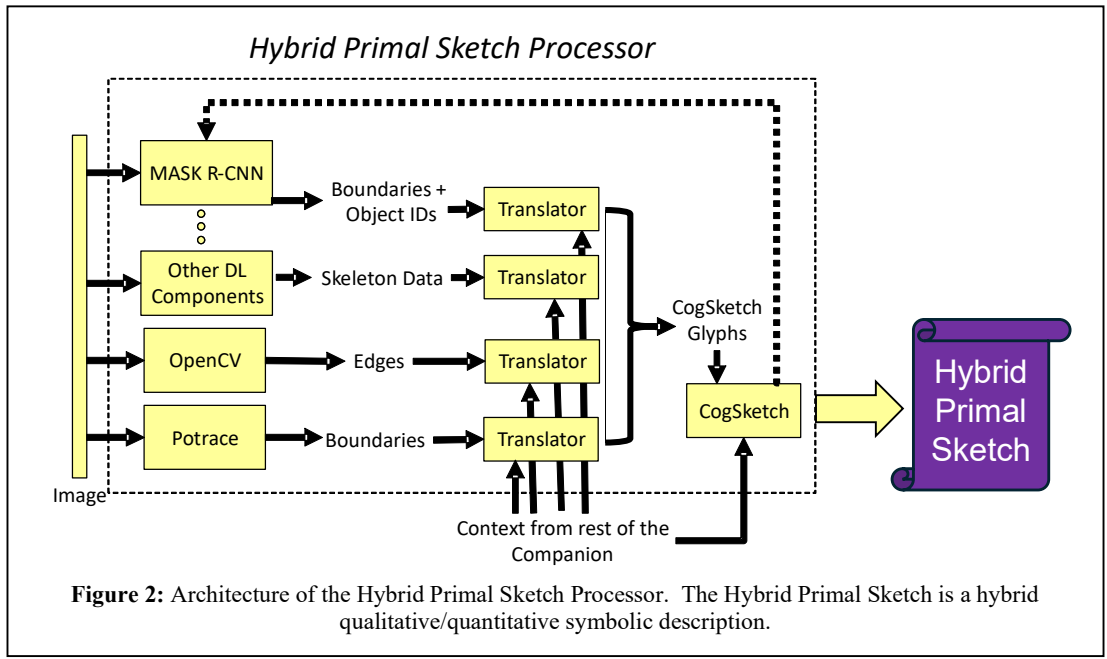

**Figure 2:** Architecture of the Hybrid Primal Sketch Processor. The Hybrid Primal Sketch is a hybrid qualitative/quantitative symbolic description.

---

[1] See Palmer (1999) for an excellent discussion.

[2] In our view, symbolic representations can and do include metric properties, e.g. coordinates, lengths, etc. See Forbus (2019), Chapter 14 for details, including evidence for psychological and neural plausibility.

Thus the Hybrid Primal Sketch consists of an ensemble of visual components (see Figure 2). But what is the common currency that these components produce, which serves as an input to the more inspectable levels of processing? Our hypothesis is that the notion of *glyph* in open-domain sketch understanding (e.g. CogSketch; Forbus et al. 2011) provides that output format. In sketch understanding, a glyph is a visual entity consisting of digital ink and a conceptual label, produced via someone drawing. In some datasets, visual stimuli are encoded in the vector graphics format SVG, which can be directly imported into CogSketch. For bitmaps of drawings, there are algorithms which can produce such vector representations automatically (i.e. Potrace, Section 2.1), and hence can be treated as glyphs by CogSketch. For natural images, the bounding boxes or masks produced by recognizers (e.g. MASK-RCNN, Section 2.1) plus the label they provide can be treated as if they were glyphs. Thus the elements of the ensemble produce glyphs for CogSketch, which then analyzes them in the same ways that it analyzes human-drawn sketches.

CogSketch was designed as a model of high-level human vision. Prior work has shown that it can be used to model human visual problem solving (Forbus & Lovett, 2022) and as a component in multimodal learning by reading (Chang 2016) and modeling conceptual change (Friedman, et al. 2018). The use of analogy in Sketch Worksheets, a deployed educational software system (Forbus et al. 2017a; Forbus et al. 2018) provides additional evidence that the representations it produces make sense to people and can be used robustly with a broad range of inputs. The results in subsequent sections provide more evidence for this claim, in terms of learning visual concepts from images (and Kinect video).

Given an image and a task context, which algorithms should be used? Currently this is a procedural parameter set by the experimenter, who sets up a pipeline custom-built for that experiment, using the capabilities of the Companion architecture to streamline the process (i.e. HTN plans provide a form of scripting, and Worker agents enables parallel processing to speed up encoding). We are moving to automatic control of processing by the Companion itself, as discussed in future work.

## 4. Experiments with the Hybrid Primal Sketch

We briefly summarize some prior experiments and describe an experiment in progress, to provide evidence for the utility of this framework. We start with sketch recognition, then turn to visual relationship detection, to illustrate its capabilities for handling images. We summarize experiments using Kinect video, to show how these ideas can be extended beyond single images. Finally, we discuss an experiment in progress on diagram understanding, since it provides a testbed for exploring the roles of conceptual understanding in scene understanding.

**4.1 Sketch Recognition**

The classic analogical learning pipeline works as follows: Encode the stimuli to be learned, and add it to a SAGE generalization pool for that concept. At any time, test items can be tried with MAC/FAC against the union of the generalization pools, with the classification of the test item being the label associated with the pool the top retrieval came from. Running this pipeline on the classic MNIST dataset (LeCun et al. 1998) provides an illustration. Encoding was performed by resizing the original image to be below 300 pixels, then blurred and filtered to black and white. Potrace and Zhang-Suen's thinning algorithm were used to generate SVG for CogSketch. CogSketch broke the SVG down into edges and edge cycles, producing a geon representation (Biederman, 1987), and computed a variety of properties for segments and geons (see (Chen et al 2019) for details. We compared our performance to LeNet-5, the classic baseline. To test data efficiency, we started with many fewer training samples[3]. Table 1 illustrates the results. We note that with fewer than 1,000 training examples, LeNet-5 performs at chance, thereby demonstrating the data-efficiency of analogical learning. Moreover, while LeNet-5 achieves quite high accuracy, it requires 20 passes through the entire set of training data, whereas Analogy requires only one. We note that adding more examples beyond 500 did not substantially improve the analogy results, so they asymptote at a lower accuracy than LeNet-5 achieves. There are two ways that the analogy results might be improved. There is no guarantee that the encoding strategy we used is optimal, and so enabling the system to do its own search for better encodings is an interesting possibility. The other way would be to use near-miss learning (McLure et al. 2015), since confusability can be an issue (see Chen et al. (2019) for details).

| Method | Training Size (per digit) | Accuracy |
|---|---|---|
| Analogy | 10 x 1 epoch | 54.9% |
| Analogy | 100 x 1 epoch | 76.24 |
| Analogy | 500 x 1 epoch | 85.03% |
| LeNet-5 | 1,500 x 20 epochs | 98.3% |
| LeNet-5 | 6,000 x 20 epochs | 99.2% |

Table 1: MNIST results.

It has long been observed that human visual representations seem to involve hierarchical structure (e.g. Marr 1982; Palmer 1999). One reason why hierarchical representations are used is to manage complexity of processing – the SME algorithm is $O(n^2 \log(n))$, so there is a cost to larger representations. Consequently, we have developed a novel extension to analogical learning, *Part-based Hierarchical Analogical Learning* (PHAL), that uses hierarchical structure to improve recognition. The encoding process involves decomposing a sketch into multiple levels of description. It starts with the outer edge cycle and moves inward (Figure 3). The decomposition process stops at level three because there are rarely more levels in hand-drawn sketches. Instead of one generalization pool per concept, there are three, one per level. Encoding involves decomposing a stimulus into edge cycles up to three levels, using CogSketch to encode the shapes and relationships between parts, and adding them to the generalization pool for that concept and level[4]. Classification of a novel stimulus starts with the same encoding and decomposition process to construct descriptions at all three levels. The classification process starts with Level 1 (coarsest), picking the K concepts whose average scores are the highest. Then, only within the K concepts found for the coarsest level, repeat at Level 2, keeping the top Q categories. Finally, within the Q concepts found for Level 2, use MAC/FAC to determine the top V candidates at Level 3. Evidence from all three levels is combined, along with a bias that each part probe should retrieve a distinct item from its gpool (i.e. it is preferable that the left wing and right wing of an aircraft match to different generalizations or outliers).

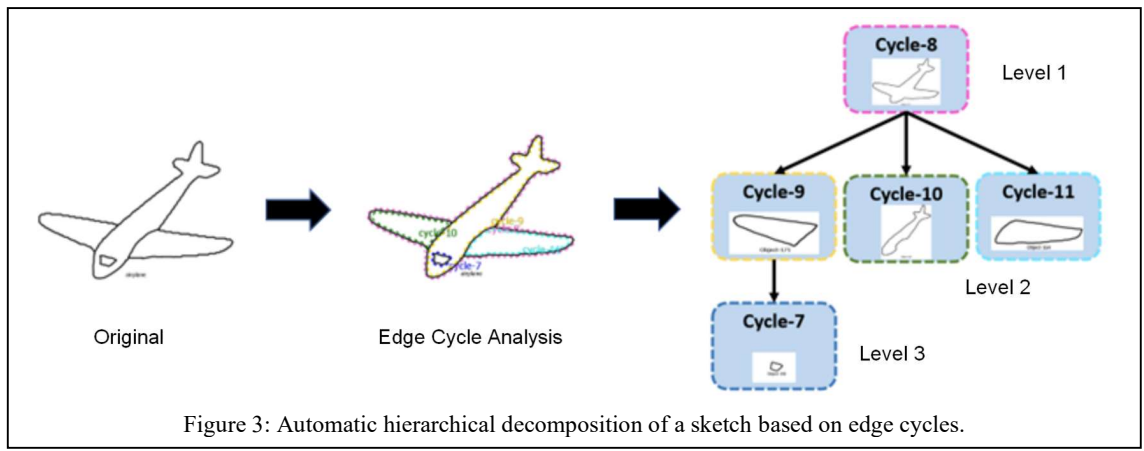

Figure 3: Automatic hierarchical decomposition of a sketch based on edge cycles.

---

[3] MNIST has 60,000 training images and 10,000 test images. Some deep learning systems generate more training data by resizing, resampling, and other manipulations.

PHAL has been tested on two datasets to date (Chen et al. 2023). The first is the TU Berlin dataset (Eitz et al. 2012), consisting of 20,000 hand drawn sketches, with 80 sketches per category and covering 250 categories. PHAL achieves 69.85% accuracy, which is competitive compared to baselines. The only systems which outperform it either do data augmentation to produce more training data or involve massive pre-training before fine-tuning on the TU Berlin dataset itself. By contrast, PHAL only uses the training data once, versus multiple epochs for the deep learning systems. On this dataset we used 10 CPUs to encode sketches and just one to perform the hierarchical analogical learning – No GPUs or TPUs required, thereby demonstrating training efficiency as well.

To focus further on data-efficiency, we developed the Coloring Book Object dataset (Chen et al. 2019). It consists of images from open-license children's coloring books, with only ten examples for each of 19 categories (animals and everyday objects). Being from coloring books, the images cover a wide range – some of the animals have hats, for example, and are drawn in a variety of styles and poses. We reproduce the results from Chen et al. (2023) in Table 2. The CNN was using the LeNet architecture, and both it and ResNet50, when trained from scratch, were at chance. Pretrained ResNet50 was better than chance, but was beaten by traditional analogical learning. PHAL does best of all, handily beating traditional analogy. This provides additional evidence for data-efficient learning via analogy, especially when exploiting hierarchical structure.

| Method | Accuracy |
|---|---|
| CNN | 5.26% |
| ResNet50 | 10.53% |
| ResNet50 (pretrained) | 21.05% |
| Analogy | 29.47% |
| PHAL | 37.37% |

Table 2: Results on CBO dataset

### 4.2 Visual Relationship Detection

---

[4] A subtlety involves texture, e.g. the pepperoni on a pizza or the patterns on the back of a turtle. As explained in Chen et al. (2023), the edges and edge cycles within a description are added to a special generalization pool, with those assimilated into a generalization being replaced by a summary of that generalization. Thus texture is also detected via analogy.

Sketches are of course simplified compared to natural images. How does the HPS approach fare on natural images? To explore this, we used a standard Visual Relationship Detection dataset (Lu et al. 2016), which consists of 5,000 images involving 100 object categories and 70 predicates. As described in (Chen & Forbus, 2021), given ground truth bounding boxes and categories (the PREDT task), analogical learning over HPS representations had 52.38 for both recall@50 and recall@100, putting it in the middle of the pack compared to prior systems, and hence competitive. We note that, as usual, the analogical learner only examined each training item once, while the deep learning systems required between seven and 30 epochs, thereby demonstrating training efficiency. (As before, the analogy learner only used CPU machines, whereas the deep learning systems use GPUs or TPUs.) While we have not been able to find recall@1 for other systems, the analogy learner had 32.26 for recall@1[5]. We also used analogical learning in the RELDT task, using off-the-shelf components (e.g. Faster-RCNN), see Chen & Forbus (2021) for details. Again our results were competitive, with recall = 16.12 @50, 18.41 @ 100, and 8.91 @1 (see Chen & Forbus, 2021 for details).

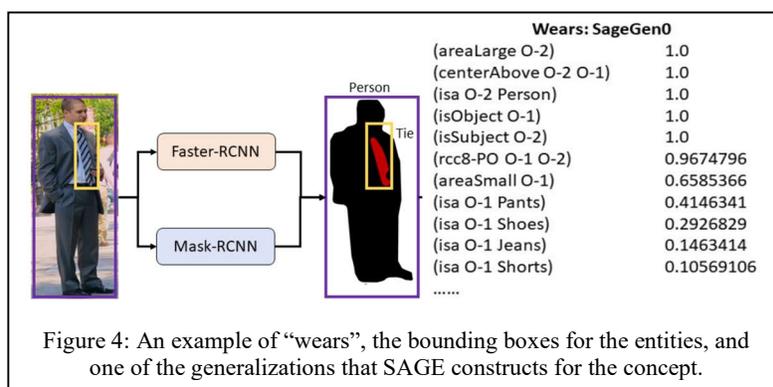

Figure 4: An example of "wears", the bounding boxes for the entities, and one of the generalizations that SAGE constructs for the concept.

In addition to being data and training efficient, analogical learning produces inspectable models. Because the visual relationships computed by CogSketch are inspired by human vision, the generalizations produced tend to be interpretable. Figure 4 shows an example of an image illustrating the "wears" relation, the bounding boxes computed for it, and one of the generalizations that SAGE constructs from the training data. (A generalization pool is a disjunctive concept description, consisting of different generalizations and outliers.) For example, in this generalization that which is being worn overlaps the person, is small compared to the size of the person, and is most likely pants, shoes, jeans, or shorts. This illustrates how the combination of the symbolic relational descriptions and probabilities produced by SAGE are inspectable.

---

[5] The problem with @N measures is that they don't provide insight to what is most commonly the long-term desired behavior, i.e. returning the right answer, or @1. We suggest that @1 measures should be reported whenever @N where N > 1 is used, to better gauge progress.

### 4.3 Human Action Recognition

Seeing images in isolation is a simplification of vision, with video being a closer approximation since organisms exist in dynamic environments. Here we examine learning from Kinect video, a special case focused on human actions. Kinect sensors produce skeleton descriptions of human motion, based on data from depth cameras, with 3D coordinates at every joint. We developed a pipeline that generates qualitative representations of such video in the form of *video sketch graphs*, a sequence of snapshots, akin to panels in a comic strip. Each snapshot describes the motion that is happening over some interval of time corresponding to a qualitative state. This temporal decomposition is accomplished with QSRLib (Gatsoulis et al. 2016), a library of qualitative spatial relations and calculi that takes streams of coordinates in, and segments them based on changes in the qualitative spatial relationships being tracked. The original 3D coordinates are projected into a front and side view, over which qualitative spatial relationships are computed for temporal segmentation.

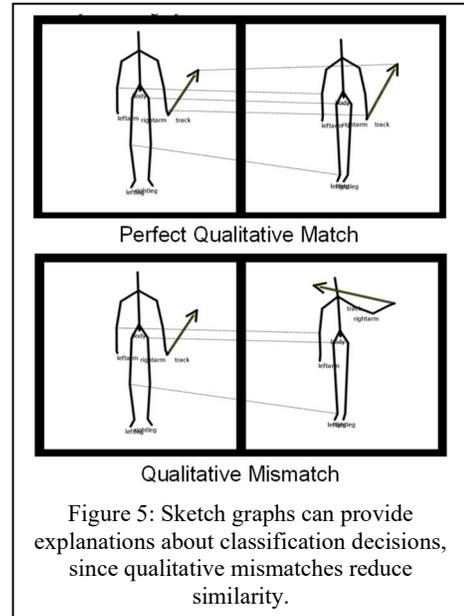

Figure 5: Sketch graphs can provide explanations about classification decisions, since qualitative mismatches reduce similarity.

There are multiple data sets of videos of simple human actions, like swiping left, clapping, throwing, jogging, walking, etc. Here we used Florence 3D Action (Seidenari et al. 2013) and UTD-MHAD (Chen et al. 2015), and UTKinect-Action3D (Xia et al. 2012) datasets. As reported in (Chen & Forbus, 2018), this approach was competitive with others, but did not perform as well with actions involving substantial noise. However, unlike the other approaches, ours provides inspectable models that can be used to explain a system's answers, as Figure 5 illustrates.

## 5. In Progress: Understanding Diagrams

Learning from instructional media, like text with diagrams, is an important task for bootstrapping AI systems. This involves understanding conventions used for depicting things in diagrams and sketches. For example, the area inside a circle intended to depict the Earth is solid, while the area inside a circle depicting an orbit is not. Lockwood et al. (2008) casts this as the problem of *conceptual segmentation*, i.e. assigning conceptual interpretations to regions and edges within a diagram or sketch. We are currently building on this approach as a further way to extend the HPS in the direction of broader conceptual interpretations, beyond just object recognition or simple visual relationship detection. We are using as a testbed the AI2 Diagrams dataset (Kembhavi et al. 2016) which focuses on types of diagrams commonly used in science educational materials.

The first challenge is to generate an initial visual segmentation as the starting point for conceptual segmentation. Here we are relying on Meta's SAM (Kirillov et al. 2023), which is capable of carving up diagrams into useful intermediate regions with reasonable accuracy. Figure 6 illustrates.

Conceptual segmentation involves determining what regions are and how they relate to each other[6], using a process of abductive reasoning. Abduction can be facilitated by having appropriate intermediate relationships. Based on its success in handling sketches in STEM domains (Forbus et al. 2017a), we think CogSketch has a set of distinctions particularly suited for diagrams. CogSketch has three kinds of glyphs: (1) *Entity glyphs* represent objects or categories, like fish or butterflies. All of the examples used in this paper so far are entity glyphs. (2) *Relation glyphs* represent binary relationships between other glyphs, such as that one animal type consumes another. Typically these are drawn by arrows but need not be. CogSketch uses proximity to conjecture which entities might be the two arguments to the relation. (3) *Annotation glyphs* represent additional information about another glyph, e.g. the flow rate of $CO_2$ between the atmosphere and ocean or a direction of motion. Figure 6 shows an example of a diagram from the dataset, the segmentation found by SAM, and the use of proximity to connect a relation glyph (i.e. one of the arrows) to entity glyphs. We are using off-the-shelf OCR to detect text labels. Visual overlap between a SAM region (or set of regions) and an OCR-detected region provides an explanation of those original SAM regions as a text label. To what does the label apply? Again, visual proximity is a common heuristic. Note that all three types of glyphs can be labeled, although relation glyphs and annotation glyphs often aren't, with conceptual context being needed to disambiguate them. For example, an arrow in a life cycle diagram often represents sequentiality in stages, whereas an arrow in a food web diagram often represents a consumption relationship.

---

[6] It can also involve determining that some regions do not have relevant conceptual interpretations, as some diagrams include decorations intended to hold reader interest.

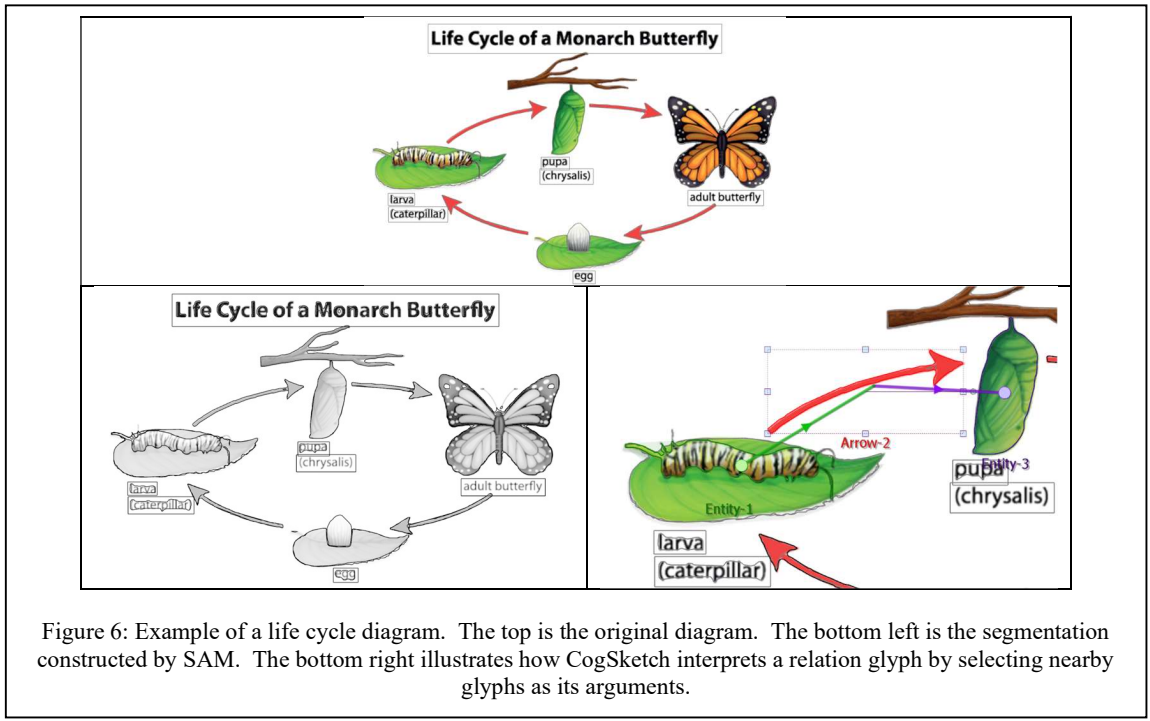

Figure 6: Example of a life cycle diagram. The top is the original diagram. The bottom left is the segmentation constructed by SAM. The bottom right illustrates how CogSketch interprets a relation glyph by selecting nearby glyphs as its arguments.

In Lockwood et al. (2008), conceptual segmentation operated over sketches drawn using CogSketch, so each glyph represented an intended visual entity. Automatic segmentation invariably has errors, so merging and splitting regions is needed. A tradeoff we will explore involves the relative roles of conceptual and visual constraints in the interpretation process. The top-down expectations imposed by text captions (e.g. "Life Cycle of…" in Figure 6) are quite strong, for instance. We are expanding our ontology with more concepts for depiction (e.g. visual containers to aggregate groups or display an entity in its setting), and setting up visual routines to interpret novel diagrams.

## 6. Conclusions and Future Work

Given the complexity of perception, finding the appropriate intermediate representations is crucial. This paper argues that CogSketch's glyph representation provides a notion of visual entity that is useful for perception beyond just sketch understanding. The ability to decompose shapes within a glyph, and represent relationships between glyphs, enables analogical learning to be used for recognition of entities and properties of scenes.

We plan three lines of future work. The first is to expand the exploration of diagram understanding, since it stress-tests the integration of conceptual and visual knowledge. The second is to make the ensemble operations more automatic. Like visual routines (Ullman 1984) and spatial routines (Lovett, 2012), we believe there is learnable procedural knowledge used to guide the application of ensemble operations to new situations and tasks. Uncovering the nature of these orienting computations is high on our agenda. Finally, we plan to continue to expand the components in the ensemble to tackle more complex problems involving natural imagery and video.

## Acknowledgements

This research was supported by the Machine Learning, Reasoning, and Intelligence Program of the US Office of Naval Research, Grant N00014-23-1-2294.